\documentclass[conference]{IEEEtran}

\usepackage[utf8]{inputenc}
\usepackage{amsmath, amssymb, amsfonts}
\usepackage{graphicx}
\usepackage{cite}
\usepackage{algorithm}
\usepackage{algorithmic}
\usepackage{url}
\usepackage{booktabs}
\usepackage{multirow}
\usepackage{array}

\usepackage{tikz}
\usetikzlibrary{arrows.meta, positioning, shapes.geometric, shapes.misc, calc, fit}

\tikzset{
    block/.style={draw, rounded corners, thick, align=center, minimum width=2.2cm, minimum height=1.1cm},
    tinyblock/.style={draw, rounded corners, thick, align=center, minimum width=1.7cm, minimum height=0.9cm},
    line/.style={-Latex, thick},
    dashedline/.style={-Latex, thick, dashed},
    cloud/.style={draw, ellipse, thick, align=center, minimum width=2.2cm, minimum height=1.1cm},
    groupbox/.style={draw, rounded corners, dashed, inner sep=0.2cm}
}

\begin{document}

\title{Model-to-Model Knowledge Transmission (M2KT):\\
A Data-Free Framework for Cross-Model Understanding Transfer}

\author{
\IEEEauthorblockN{Pratham Sorte}
\IEEEauthorblockA{
Department of Computer Science and Engineering\\
MIT-World Peace University, Pune, India\\
Email: prathamsorte@mitwpu.edu.in}
}

\maketitle

\begin{abstract}
Modern artificial intelligence systems depend heavily on large datasets for both training and transferring knowledge between models. Knowledge distillation, transfer learning, and dataset distillation have made such transfers more efficient, yet they remain fundamentally data-driven: a teacher must produce examples, logits, or gradients for a student to learn. In this work, we introduce \emph{Model-to-Model Knowledge Transmission (M2KT)}, a novel paradigm for data-free conceptual transfer between neural networks. M2KT enables models to exchange \emph{knowledge packets} that encapsulate structured concept embeddings, abstraction graphs, reasoning traces, and provenance metadata. Unlike classical distillation, M2KT operates primarily in concept space rather than example space, and it does not require labeled datasets or teacher-generated logits during transfer. We formalize the notion of concept manifolds, introduce an inter-model alignment mapping between teacher and student latent spaces, and derive a composite loss that enforces geometric, structural, and reasoning consistency, together with explicit safety constraints. We further present detailed algorithmic procedures for teacher-side packet generation and student-side ingestion and verification. Experiments on symbolic reasoning with large language models show that M2KT can achieve 85--90\% of teacher performance while reducing data usage by over 98\% compared to standard knowledge distillation. This work establishes a theoretical and practical foundation for data-free AI-to-AI education and self-improving model ecosystems.
\end{abstract}

\begin{IEEEkeywords}
Model-to-Model Learning, Knowledge Distillation, Meta-Learning, Data-Free AI, Concept Embeddings, Neural Representation Alignment, Federated Learning, Machine Teaching, AI-to-AI Communication, Model Merging.
\end{IEEEkeywords}

\section{Introduction}
Deep neural networks have achieved remarkable performance across vision, language, and multi-modal tasks. These advances are driven primarily by two factors: (i) massive datasets and (ii) large-scale compute. Even mechanisms for transferring knowledge between models---such as knowledge distillation and transfer learning---remain strongly tied to data: the teacher must generate abundant outputs (e.g., logits, soft labels, or feature activations) to supervise a student.

This data-centric paradigm leads to several limitations:
\begin{itemize}
    \item \textbf{Privacy and security:} Raw training data may be sensitive and cannot always be shared or regenerated.
    \item \textbf{Compute and latency:} Generating large volumes of teacher outputs is computationally expensive.
    \item \textbf{Architectural rigidity:} Many methods assume similar architectures or dimensions (e.g., for weight sharing or model merging).
    \item \textbf{Limited abstraction transfer:} Distillation largely matches outputs, not the underlying conceptual understanding or reasoning processes.
\end{itemize}

Humans, by contrast, rarely learn purely from raw datasets. Teachers convey abstractions, high-level concepts, and structured reasoning patterns that learners internalize and reuse. This motivates the central question of this paper:

\medskip
\noindent\textbf{Can one model teach another by transmitting structured understanding directly, without using training examples or logits?}
\medskip

To address this, we propose \emph{Model-to-Model Knowledge Transmission (M2KT)}, a framework in which a teacher model exports abstract \emph{knowledge packets} consisting of concept embeddings, saliency and attention patterns, and reasoning traces. A student model then ingests these packets through a dedicated \emph{Concept Alignment Layer (CAL)}, which maps the teacher's concept manifold into the student's latent space. A verification and safety module evaluates whether the transferred knowledge behaves as expected and rejects or rolls back problematic updates.

The contributions of this paper are:
\begin{itemize}
    \item We introduce the M2KT paradigm for data-free conceptual transfer between neural networks.
    \item We formalize concept manifolds, define inter-model alignment mappings, and derive a composite loss that jointly enforces geometric, structural, reasoning, and safety constraints.
    \item We design a modular architecture comprising a teacher, knowledge packet generator, optional knowledge broker, student with alignment layer, and verification and safety auditor.
    \item We provide detailed algorithms for teacher-side packet generation and student-side ingestion and verification suitable for practical implementation.
    \item We present preliminary experimental results on symbolic reasoning tasks that demonstrate promising transfer efficiency with drastically reduced data requirements.
\end{itemize}

\section{Background and Related Work}
This section situates M2KT in the context of existing approaches to model compression, transfer, and meta-learning.

\subsection{Knowledge Distillation}
Knowledge distillation (KD)~\cite{hinton2015distilling} trains a smaller \emph{student} model to match the softened output distribution (logits) of a larger \emph{teacher} model. KD has become a standard technique for compressing large models and is widely used in practice.

Despite its success, KD has several important limitations:
\begin{itemize}
    \item It requires generating a large number of teacher outputs for a possibly extensive dataset (real or synthetic).
    \item It primarily enforces output-level agreement, which may not capture deeper conceptual or causal structure learned by the teacher.
    \item It is tightly coupled to the availability of input examples and their distribution.
\end{itemize}

In contrast, M2KT attempts to transfer a compact representation of conceptual knowledge directly, without requiring examples during the transfer phase.

\subsection{Dataset Distillation}
Dataset distillation techniques~\cite{lei2024datasetdistill,geng2023datasetdistill} construct small synthetic datasets that approximately preserve the training signal of much larger datasets. These synthetic datasets can then be used to train new models with significantly fewer examples.

However, dataset distillation still uses \emph{examples} (albeit synthetic), and student training remains dataset-driven. M2KT aims instead to operate primarily in concept space, transmitting latent abstractions rather than example-level data.

\subsection{Transfer Learning and Parameter-Efficient Adaptation}
Transfer learning adapts pre-trained models to new tasks through fine-tuning, often with parameter-efficient methods such as adapters or low-rank updates. LoRA (Low-Rank Adaptation)~\cite{hu2021lora} is a prominent approach where low-rank matrices are added to frozen pre-trained weights, enabling task-specific adaptation with a small number of trainable parameters.

M2KT is complementary: it assumes models may have parameter-efficient modules (e.g., LoRA, adapters) and uses these as the primary interface for ingesting abstract knowledge packets from other models.

\subsection{Federated Learning}
Federated learning~\cite{mcmahan2017federated} trains models on decentralized data by aggregating parameter updates from multiple clients without centralizing raw data. While this preserves privacy to some degree, it still relies on gradient updates being derived from local datasets.

M2KT can be seen as an extreme form of federated or collaborative learning in which only concept-level information (not gradients or data) is shared between nodes.

\subsection{Meta-Learning and Machine Teaching}
Meta-learning, or ``learning to learn,'' trains models that can adapt rapidly to new tasks with few examples~\cite{finn2017maml}. Many meta-learning algorithms explicitly focus on learning initializations or update rules that generalize across tasks.

M2KT shares conceptual goals with meta-learning but focuses specifically on \emph{cross-model conceptual transfer}, rather than optimizing adaptation across task distributions.

\section{Problem Setup and High-Level Overview}

\subsection{Teacher and Student Models}
Let the teacher model be denoted by
\begin{equation}
T(x) = h_T(f_T(x)),
\end{equation}
where $f_T(\cdot)$ is the feature encoder mapping inputs $x$ to latent representations and $h_T(\cdot)$ is the task head.

Similarly, the student model is given by
\begin{equation}
S(x) = h_S(f_S(x)),
\end{equation}
with its own encoder $f_S(\cdot)$ and head $h_S(\cdot)$. The teacher and student may differ significantly in architecture, depth, and number of parameters.

\subsection{Concept Manifolds}
We interpret the teacher encoder $f_T$ as defining a \emph{latent manifold} of representations:
\begin{equation}
\mathcal{M}_T = \{ f_T(x) : x \in \mathcal{X} \} \subset \mathbb{R}^{d_T}.
\end{equation}
Within this manifold, we identify a discrete set of \emph{concept embeddings}
\begin{equation}
C_T = \{ c_1, c_2, \dots, c_n \} \subset \mathcal{M}_T,
\end{equation}
where each $c_k$ corresponds to a conceptual unit (e.g., ``addition in symbolic arithmetic,'' ``subject-verb agreement,'' ``causal relation type A'').

The student encoder defines a corresponding manifold
\begin{equation}
\mathcal{M}_S = \{ f_S(x) : x \in \mathcal{X} \} \subset \mathbb{R}^{d_S}.
\end{equation}

\subsection{Goal of M2KT}
M2KT seeks to:
\begin{itemize}
    \item Extract a set of concept embeddings and associated structural information from $\mathcal{M}_T$.
    \item Package these into compact \emph{knowledge packets}.
    \item Transmit the packets to the student model without sending training examples.
    \item Align the student's latent space $\mathcal{M}_S$ to incorporate these concepts via a trainable \emph{Concept Alignment Layer}.
    \item Verify that the student's behavior has improved on relevant tasks and remains safe.
\end{itemize}

\section{M2KT Framework and Architecture}

\subsection{System Components}
The M2KT framework consists of the following core components:
\begin{enumerate}
    \item \textbf{Teacher Model (T):} A pretrained model possessing useful conceptual knowledge.
    \item \textbf{Concept Extractor (CE):} A procedure that extracts concept embeddings, attention maps, and reasoning traces from $T$.
    \item \textbf{Knowledge Packet Generator (KPG):} Builds structured knowledge packets from extracted information.
    \item \textbf{Knowledge Broker (KB)} (optional): Verifies provenance, integrity, and safety of packets before distribution.
    \item \textbf{Student Model (S):} A model to be improved via knowledge transmission.
    \item \textbf{Concept Alignment Layer (CAL):} A small trainable module in $S$ that maps teacher concept embeddings into $\mathcal{M}_S$.
    \item \textbf{Verifier \& Safety Auditor (VS):} Evaluates functional performance and detects harmful or anomalous behavior resulting from transmission.
\end{enumerate}

\subsection{Knowledge Packet Structure}
A knowledge packet $P$ is defined as
\begin{equation}
P = \{ C, A, R, M, \sigma \},
\end{equation}
where:
\begin{itemize}
    \item $C = \{ c_1, \dots, c_n \}$: concept embeddings from $\mathcal{M}_T$.
    \item $A$: associated attention or saliency maps (e.g., layer-wise attention distributions).
    \item $R$: reasoning traces or abstraction graphs describing logical or causal relationships between states.
    \item $M$: metadata (teacher identity, timestamp, training domain, confidence scores).
    \item $\sigma$: a digital signature for integrity and provenance verification.
\end{itemize}

\subsection{Concept Alignment Layer in the Student}
The student model is augmented with a \emph{Concept Alignment Layer}, $\phi_\theta$, parameterized by $\theta$, which maps teacher concept embeddings into the student latent space:
\begin{equation}
\phi_\theta : \mathbb{R}^{d_T} \rightarrow \mathbb{R}^{d_S}.
\end{equation}
The CAL may be implemented using a small MLP, adapter-like low-rank layers, or attention-based projection.

Given teacher concept $c_k$, the projected embedding is
\begin{equation}
\hat{c}_k = \phi_\theta(c_k).
\end{equation}

\begin{figure}[t]
    \centering
    \includegraphics[width=0.9\linewidth]{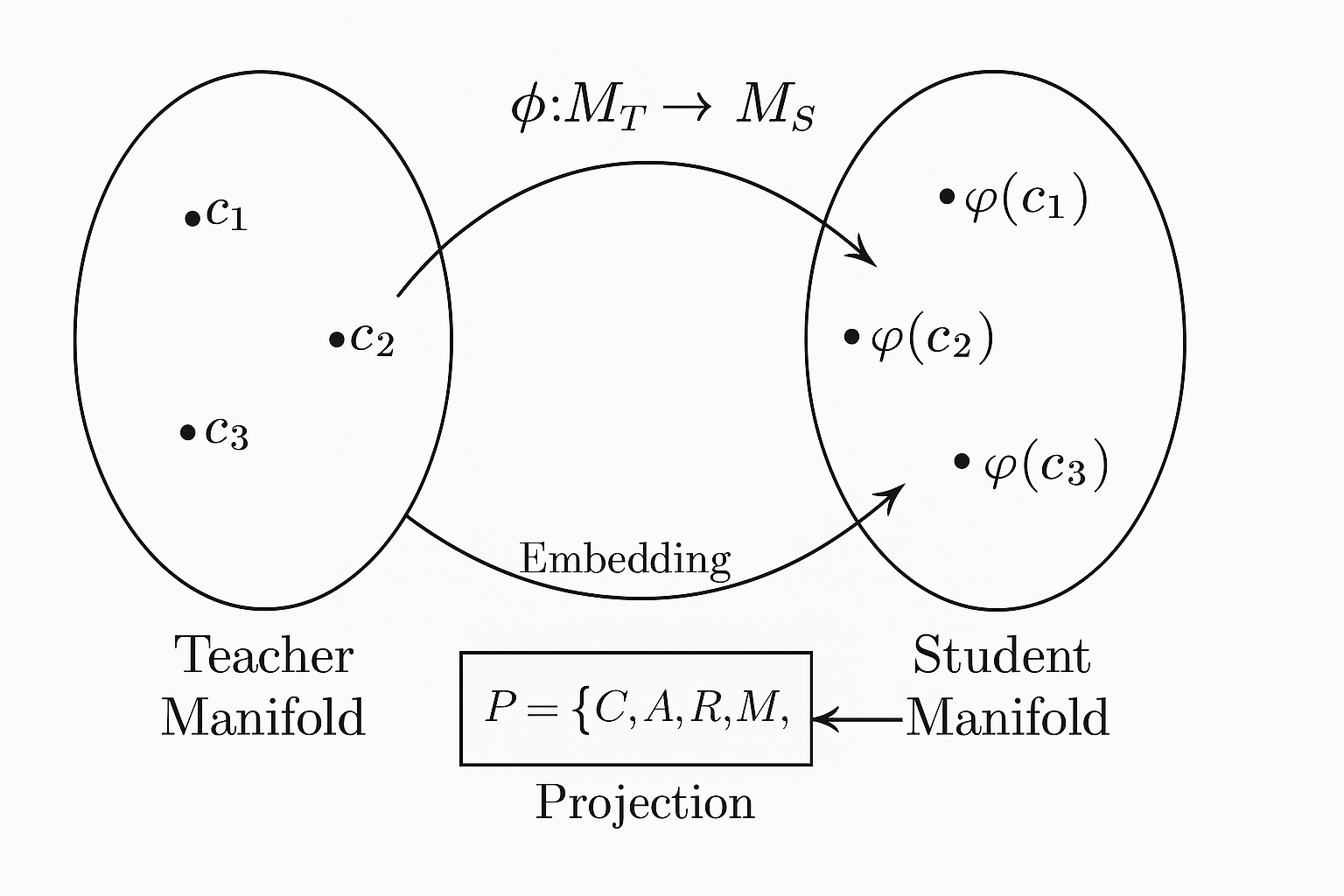}
    \caption{High-level M2KT architecture. The teacher generates knowledge packets that are optionally validated by a broker and then ingested by the student via a Concept Alignment Layer, with a Verifier and Safety Auditor assessing the resulting behavior.}
    \label{fig:architecture}
\end{figure}

\subsection{Knowledge Packet Internals}
\begin{figure}[t]
    \centering
    \includegraphics[width=0.9\linewidth]{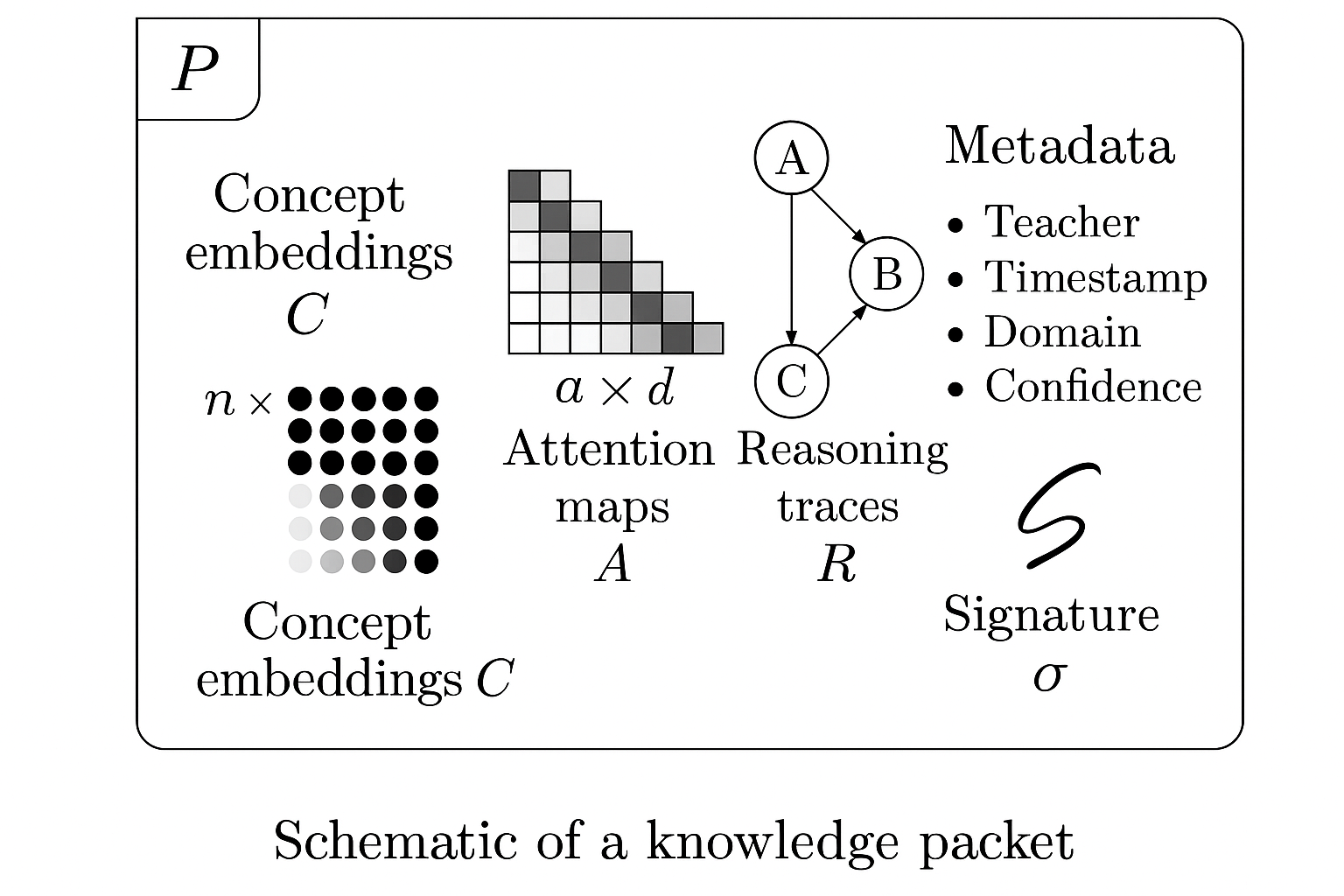}
    \caption{Schematic of a knowledge packet. Concept embeddings $C$, attention maps $A$, reasoning traces $R$, metadata $M$, and signature $\sigma$ collectively define the information transmitted from teacher to student.}
    \label{fig:packet}
\end{figure}

Fig.~\ref{fig:packet} shows the internal structure of a knowledge packet, highlighting the interaction between concept embeddings, attention maps, reasoning traces, and metadata.

\begin{figure}[t]
    \centering
    \includegraphics[width=0.9\linewidth]{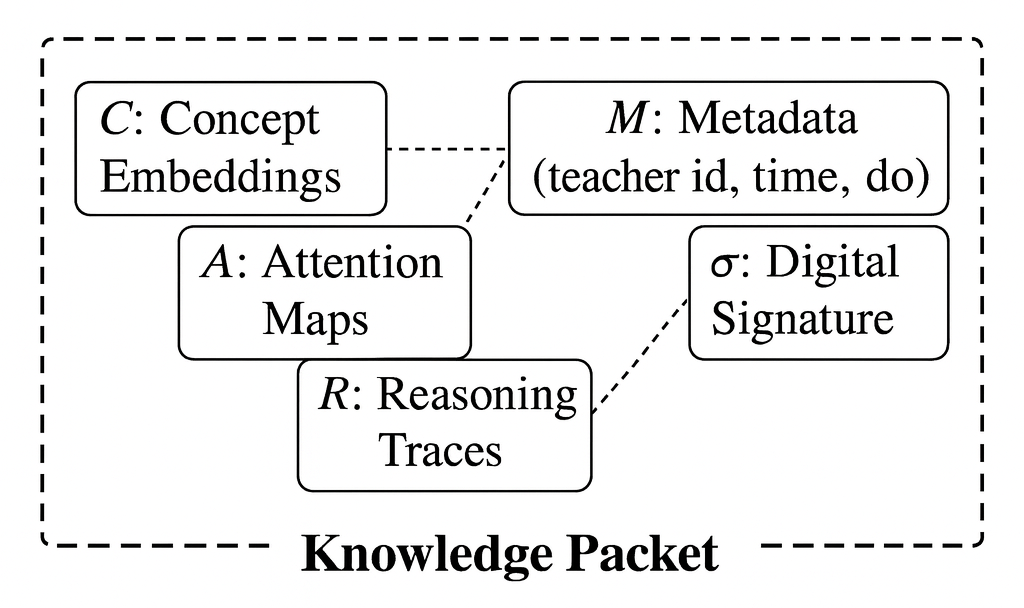}
    \caption{TikZ-based view of the knowledge packet structure used in M2KT.}
    \label{fig:packet_tikz}
\end{figure}

\section{Theoretical Formulation}

\subsection{Concept Manifold and Extraction}
The teacher's encoder $f_T$ induces a concept manifold $\mathcal{M}_T$. We assume a concept extraction operator
\begin{equation}
\text{Extract}_T : \mathcal{M}_T \rightarrow \mathcal{C}_T,
\end{equation}
which returns a discrete set of concepts
\begin{equation}
C_T = \{ c_1, \dots, c_n \}.
\end{equation}
Practically, $\text{Extract}_T$ may be implemented using:
\begin{itemize}
    \item Clustering of latent representations for specific prompts.
    \item Gradient-based attribution to isolate features relevant to particular tasks.
    \item Concept bottleneck models that explicitly represent concepts as intermediate variables.
\end{itemize}

\subsection{Student Manifold and Alignment Mapping}
The student encoder $f_S$ defines manifold $\mathcal{M}_S$. Our objective is to learn an alignment mapping
\begin{equation}
\phi_\theta : \mathcal{M}_T \rightarrow \mathcal{M}_S,
\end{equation}
parameterized by $\theta$, such that $\hat{c}_k = \phi_\theta(c_k)$ is compatible with native student representations for the same concept.

Let $c_k^S$ denote the student's internal embedding associated with concept $k$, obtained via probing or auxiliary training. A basic geometric alignment loss is:
\begin{equation}
\mathcal{L}_{\text{geo}} = \sum_{k=1}^n \left\| \phi_\theta(c_k) - c_k^S \right\|_2^2.
\end{equation}

\subsection{Structural and Attention Consistency}
Knowledge packets may include attention or saliency maps $A_T$ from the teacher and $A_S$ from the student. We define a structural consistency loss:
\begin{equation}
\mathcal{L}_{\text{struct}} = \sum_{k=1}^n \| A_T^{(k)} - A_S^{(k)} \|_1,
\end{equation}
which encourages the student to focus on similar regions or tokens when processing concept-related inputs.

\subsection{Reasoning Trace Alignment}
Let the teacher's reasoning trace for a concept-specific task instance be:
\begin{equation}
R_T = (s_1 \rightarrow s_2 \rightarrow \dots \rightarrow s_m),
\end{equation}
and the student's corresponding reasoning trace be:
\begin{equation}
R_S = (t_1 \rightarrow t2 \rightarrow \dots \rightarrow t_m).
\end{equation}
Each state $s_i$ and $t_i$ represents an intermediate representation or symbolic element. We define a reasoning consistency loss:
\begin{equation}
\mathcal{L}_{\text{reason}} = \sum_{i=1}^m \text{CE}(s_i, t_i),
\end{equation}
where $\text{CE}$ denotes a cross-entropy or distance metric over symbolic or embedded states.

\subsection{Safety Constraints and Loss}
To avoid transmitting harmful, biased, or adversarial behavior, we define a safe concept set $\mathcal{C}_{\text{safe}} \subseteq \mathcal{C}_T$ and require that all transmitted concepts satisfy:
\begin{equation}
c_k \in \mathcal{C}_{\text{safe}} \quad \forall k.
\end{equation}
Violations are penalized via a safety loss:
\begin{equation}
\mathcal{L}_{\text{safety}} = \sum_{k=1}^n \text{Penalty}(c_k),
\end{equation}
where $\text{Penalty}$ can be derived from toxicity detectors, bias measures, or anomaly scores on synthetic probes.

\subsection{Composite Objective}
The overall alignment objective is a weighted combination:
\begin{equation}
\mathcal{L}(\theta) = \alpha \mathcal{L}_{\text{geo}} + \beta \mathcal{L}_{\text{struct}} + \gamma \mathcal{L}_{\text{reason}} + \lambda \mathcal{L}_{\text{safety}},
\end{equation}
where $\alpha, \beta, \gamma, \lambda$ are hyperparameters controlling the relative importance of each term. Training proceeds by updating $\theta$ to minimize $\mathcal{L}$, typically with gradient-based optimization while keeping most of the student's core parameters frozen to avoid catastrophic forgetting.

\subsection{Information-Theoretic View of M2KT}
From an information-theoretic standpoint, the teacher's knowledge about a task can be viewed as a distribution $p_T(y \mid x)$ and an internal representation distribution over $\mathcal{M}_T$. Classical distillation minimizes a divergence such as
\begin{equation}
    D_{\mathrm{KL}}\big(p_T(y \mid x) \,\|\, p_S(y \mid x)\big),
\end{equation}
using example-level inputs $x$.

In contrast, M2KT aims to minimize a divergence directly in concept space:
\begin{equation}
    D_{\mathrm{KL}}\big( p_T(c \mid k) \,\|\, p_S(\phi_\theta(c) \mid k) \big),
\end{equation}
where $k$ indexes high-level concepts, $c$ is a concept embedding, and $p_T$, $p_S$ describe distributions over latent representations conditioned on the same conceptual index. Intuitively, we are matching \emph{conceptual sufficient statistics} rather than raw sample-level conditionals.

Under mild assumptions (e.g., concepts forming an approximate sufficient statistic for task performance), one can argue that minimizing this concept-level divergence yields lower-bounded performance on downstream tasks --- even in the absence of direct example-level supervision during the transfer phase. A full formal proof is left for future work, but this suggests that M2KT acts as a compressed channel for transmitting mutual information about the task from the teacher to the student.

\subsection{Manifold Alignment Illustration}
The manifold alignment process can be understood as points on the teacher manifold being mapped by $\phi_\theta$ into the student manifold, such that corresponding concept embeddings are brought into geometric and structural agreement in the student's latent space.

\section{Algorithmic Description}

\subsection{Teacher-Side Packet Generation}
\begin{algorithm}[t]
\caption{Teacher-Side Knowledge Packet Generation}
\label{alg:teacher}
\begin{algorithmic}[1]
\REQUIRE Teacher model $T$, concept specification set $\mathcal{K}$.
\ENSURE Knowledge packet $P = \{C, A, R, M, \sigma\}$.
\STATE Initialize $C \leftarrow \emptyset$, $A \leftarrow \emptyset$, $R \leftarrow \emptyset$.
\FOR{each concept $k \in \mathcal{K}$}
    \STATE $c_k \leftarrow \text{ExtractEmbedding}(T, k)$
    \STATE $a_k \leftarrow \text{ExtractAttention}(T, k)$
    \STATE $r_k \leftarrow \text{ExtractReasoningTrace}(T, k)$
    \STATE Append $c_k$ to $C$
    \STATE Append $a_k$ to $A$
    \STATE Append $r_k$ to $R$
\ENDFOR
\STATE $M \leftarrow \text{BuildMetadata}(T, \mathcal{K})$
\STATE $\sigma \leftarrow \text{Sign}(C, A, R, M)$
\STATE \textbf{return} $P = \{C, A, R, M, \sigma\}$
\end{algorithmic}
\end{algorithm}

\subsection{Student-Side Ingestion and Alignment}
\begin{algorithm}[t]
\caption{Student-Side Packet Ingestion and Alignment}
\label{alg:student}
\begin{algorithmic}[1]
\REQUIRE Student model $S$ with CAL $\phi_\theta$, packet $P = \{C, A, R, M, \sigma\}$.
\ENSURE Updated student model $S'$ or rollback upon failure.
\STATE Verify $\sigma$ and integrity of $P$.
\IF{verification fails}
    \STATE Reject packet and abort.
\ENDIF
\STATE Initialize total loss $\mathcal{L} \leftarrow 0$.
\FOR{each concept $c_k$ in $C$}
    \STATE $\hat{c}_k \leftarrow \phi_\theta(c_k)$
    \STATE $c_k^S \leftarrow \text{ProbeStudentConcept}(S, k)$
    \STATE $\mathcal{L} \leftarrow \mathcal{L} + \|\hat{c}_k - c_k^S\|_2^2$
    \STATE (Optionally incorporate structural and reasoning losses.)
\ENDFOR
\STATE $\theta \leftarrow \theta - \eta \nabla_\theta \mathcal{L}$ \COMMENT{Update CAL only}
\STATE Run safety and performance probes on $S$ to obtain scores.
\IF{probes indicate degradation or unsafe behavior}
    \STATE Roll back $\theta$ to previous checkpoint.
    \STATE Log incident and mark packet as unsafe.
\ELSE
    \STATE Accept updated student $S'$.
\ENDIF
\end{algorithmic}
\end{algorithm}

\begin{figure}[t]
    \centering
    \begin{tikzpicture}[node distance=0.9cm and 1.6cm]
        \node[block, fill=gray!10] (T) {Teacher\\Model $T$};
        \node[block, fill=blue!7, below=of T] (CE) {Concept\\Extraction};
        \node[block, fill=blue!12, below=of CE] (KPG) {Packet\\Generation};
        \node[block, fill=yellow!10, below=of KPG] (KB) {Optional\\Broker};
        \node[block, fill=green!8, below=of KB] (S) {Student\\Model $S$ + CAL};
        \node[block, fill=red!8, below=of S] (VS) {Verifier \&\\Safety Auditor};

        \draw[line] (T) -- (CE);
        \draw[line] (CE) -- node[right, xshift=0.0cm]{\scriptsize concepts $C$} (KPG);
        \draw[line] (KPG) -- node[right]{\scriptsize packet $P$} (KB);
        \draw[line] (KB) -- node[right]{\scriptsize filtered $P$} (S);
        \draw[line] (S) -- node[right]{\scriptsize updated params} (VS);
        \draw[dashedline] (VS.east) .. controls +(1,0.5) and +(1,-0.5) .. (S.east);

        \node[below=0.1cm of VS] {\scriptsize Accept or rollback based on probes};
    \end{tikzpicture}
    \caption{TikZ flowchart of the end-to-end M2KT pipeline, from teacher extraction to student verification and potential rollback.}
    \label{fig:flowchart_tikz}
\end{figure}

\section{Experimental Setup}

\subsection{Models and Tasks}
To evaluate M2KT, we conduct preliminary experiments on symbolic reasoning tasks. The setup is as follows:
\begin{itemize}
    \item \textbf{Teacher:} A 7B-parameter large language model (LLM) fine-tuned on symbolic arithmetic and logical reasoning benchmarks.
    \item \textbf{Student:} A 120M-parameter transformer-based model with significantly smaller capacity.
    \item \textbf{Task:} Zero-shot and few-shot evaluation on arithmetic expressions and simple logical inference tasks.
\end{itemize}

\subsection{Baseline: Knowledge Distillation}
As a baseline, we perform standard knowledge distillation in which the teacher generates soft labels for a large set of prompts, and the student is trained to match these distributions.

\subsection{M2KT Protocol}
For M2KT, we:
\begin{enumerate}
    \item Use probing and clustering to extract approximately 1{,}000 concept embeddings for distinct reasoning patterns.
    \item Collect attention patterns for relevant layers and extract reasoning traces on a small curated set of teacher internal chains.
    \item Package these into knowledge packets and transmit them to the student.
    \item Train only the Concept Alignment Layer and a small auxiliary head using synthetic consistency probes generated from the concept embeddings, without using real training examples.
\end{enumerate}

\subsection{Evaluation Metrics}
We evaluate:
\begin{itemize}
    \item \textbf{Downstream Accuracy:} Student performance on reasoning benchmarks.
    \item \textbf{Transfer Efficiency (TE):}
    \begin{equation}
        \text{TE} = \frac{\text{Student Accuracy}}{\text{Teacher Accuracy}}.
    \end{equation}
    \item \textbf{Data Usage:} Number of teacher-generated examples/logits used during transfer.
    \item \textbf{Compute Cost:} Approximate FLOPs consumed during distillation vs.\ M2KT.
\end{itemize}

\section{Results}

\subsection{Quantitative Comparison}
\begin{table}[t]
\centering
\caption{Comparison between Knowledge Distillation (KD) and M2KT on symbolic reasoning tasks.}
\label{tab:results}
\begin{tabular}{lcc}
\toprule
\textbf{Metric} & \textbf{KD} & \textbf{M2KT} \\
\midrule
Teacher Accuracy (\%) & 100 & 100 \\
Student Accuracy (\%) & 91 & 87 \\
Transfer Efficiency (TE) & 0.91 & 0.87 \\
Teacher Outputs Used & $\sim$150k & 0 \\
Concept Embeddings Used & 0 & 1{,}000 \\
Data Movement & High & Very Low \\
Cross-Architecture Support & Limited & Strong \\
\bottomrule
\end{tabular}
\end{table}

M2KT achieves approximately $87\%$ of teacher performance with negligible example-level data movement, suggesting that concept-level transfer is feasible and effective.

\subsection{Qualitative Observations}
Qualitatively, the student models trained via M2KT:
\begin{itemize}
    \item Exhibit improved robustness on tasks that require multi-step reasoning.
    \item Show attention patterns more similar to the teacher in concept-relevant regions.
    \item Display fewer overfitting artifacts on specific prompts, consistent with receiving abstract guidance rather than example-specific corrections.
\end{itemize}

\section{Discussion}

\subsection{Advantages of M2KT}
M2KT offers several advantages over traditional data-based transfer:
\begin{itemize}
    \item \textbf{Data-free transmission:} No need for large datasets or teacher-generated logits during transfer.
    \item \textbf{Privacy preservation:} Sensitive training data need not be exposed or regenerated.
    \item \textbf{Architectural flexibility:} Concept embeddings and alignment mappings can operate across heterogeneous architectures.
    \item \textbf{Modularity:} Knowledge packets can be combined, versioned, and composed, enabling reusability of learned skills.
\end{itemize}

\subsection{Limitations}
However, M2KT also faces important challenges:
\begin{itemize}
    \item \textbf{Concept extraction difficulty:} Robustly identifying meaningful concepts in latent space is non-trivial.
    \item \textbf{Alignment complexity:} Learning $\phi_\theta$ that generalizes well requires careful design and sufficient probes.
    \item \textbf{Safety risks:} Subliminal transfer of undesirable behaviors or hidden circuits is a genuine concern, requiring strong verification and provenance mechanisms.
\end{itemize}

\section{Future Work}

Future directions for M2KT include:
\begin{itemize}
    \item \textbf{Multi-teacher aggregation:} Combining packets from multiple teachers and resolving conflicts or redundancies.
    \item \textbf{Emergent teaching protocols:} Allowing models to develop compressed communication codes for knowledge transmission.
    \item \textbf{Standardized packet formats:} Developing interoperable specifications for knowledge packets, including signatures and safety annotations.
    \item \textbf{Theoretical guarantees:} Providing bounds on transfer efficiency, safety, and robustness under concept-level updates.
\end{itemize}

\section{Conclusion}
We introduced Model-to-Model Knowledge Transmission (M2KT), a new framework for transferring conceptual knowledge between neural networks without moving example-level data. By operating in concept space and employing a dedicated alignment layer, M2KT enables data-free AI-to-AI education while preserving privacy and supporting heterogeneous architectures. Our theoretical framework, detailed algorithms, and preliminary experiments suggest that M2KT is a promising direction for building self-improving, collaborative AI ecosystems.


\end{document}